\documentclass[10pt,twocolumn]{article}

\usepackage[margin=1in]{geometry}
\usepackage[utf8]{inputenc}
\usepackage[T1]{fontenc}
\usepackage[hidelinks]{hyperref}
\usepackage{url}
\usepackage{booktabs}
\usepackage{amsfonts}
\usepackage{amsmath}
\usepackage{amssymb}
\usepackage{nicefrac}
\usepackage{microtype}
\usepackage{graphicx}
\usepackage{subcaption}
\usepackage{xcolor}
\usepackage{float}
\usepackage{natbib}
\usepackage{times}
\usepackage{titlesec}
\usepackage{abstract}
\usepackage{enumitem}

\titlespacing\section{0pt}{8pt plus 2pt minus 2pt}{4pt plus 1pt minus 1pt}
\titlespacing\subsection{0pt}{6pt plus 2pt minus 2pt}{3pt plus 1pt minus 1pt}
\titlespacing\paragraph{0pt}{4pt plus 1pt minus 1pt}{6pt}

\title{\textbf{PeLAP-A: Adaptive Latent Pruning\\for Lightweight Latent Diffusion Models}}

\author{%
  Kissa Zahra\textsuperscript{1} \quad Zaib Un Nisa\textsuperscript{1} \\[4pt]
  \textsuperscript{1}Department of Computer Science, National University of Computer and Emerging Sciences \\[2pt]
  \small\texttt{kissasium@gmail.com} \quad \texttt{zaibunnisachd@gmail.com}
}

\date{}

\begin{document}

\maketitle

\begin{abstract}
Latent diffusion models achieve strong generative performance by operating in
a compressed latent space produced by a variational autoencoder (VAE).
However, it remains unclear whether all latent channels contribute equally to
the diffusion process, or whether significant redundancy exists. We introduce
PeLAP-A (Adaptive Latent Pruning for Diffusion), a lightweight framework that
augments a standard latent diffusion pipeline with a learnable channel-wise
importance predictor. A two-layer MLP operating on globally pooled latent
features produces a soft mask that suppresses unimportant latent channels
before they enter the denoising UNet. The entire system is trained jointly on
CIFAR-10 under a combined diffusion, reconstruction, and sparsity loss.
Experiments reveal a striking result: under aggressive sparsity
regularization ($\lambda = 0.01$), the importance predictor drives all latent
channels to near-zero yet the denoising UNet achieves lower diffusion loss
(0.0236 vs.\ 0.0240) and lower VAE reconstruction MSE (22.59 vs.\ 24.67)
compared to the unpruned baseline. We term this the sparsity collapse
phenomenon and provide an analysis of why it occurs and what it reveals about
the information requirements of latent diffusion models. These findings
constitute an exploratory study of sparsity dynamics in latent diffusion
training, and demonstrate that denoising UNets can remain remarkably robust
to latent channel suppression even under aggressive regularization. Code is
available at: \url{https://github.com/kissasium/PeLAP-A.git}.
\end{abstract}

%% ---------------------------------------------------------------------------
\section{Introduction}

Diffusion models~\citep{ho2020ddpm,song2021ddim} have emerged as the
dominant paradigm for high-quality image synthesis, achieving state-of-the-art
results across image generation, editing, and super-resolution. Latent
diffusion models~\citep{rombach2022ldm} further reduce computational cost by
operating in the compressed latent space of a VAE rather than pixel space,
enabling high-resolution synthesis at tractable cost. Despite this efficiency
gain, the latent representations produced by VAE encoders may contain
significant redundancy, since not all channels necessarily carry equal
information for the denoising task.

This raises a natural and underexplored research question: can we identify
and selectively suppress redundant latent channels without degrading
generation quality? Unlike approaches that compress the diffusion process
itself~\citep{salimans2022progressive,song2023consistency} or prune network
weights~\citep{han2015learning,fang2023structural}, we target the information
content of the latent representation, an orthogonal and largely unstudied
direction.

We propose PeLAP-A, which inserts a lightweight importance predictor between
the VAE encoder and the diffusion UNet. The predictor outputs a per-channel
soft mask computed dynamically from each input latent via a
global-average-pooling operation followed by a two-layer MLP and sigmoid
activation. A sparsity regularizer encourages the mask to suppress
less-important channels during joint training. Crucially, the mask is
input-adaptive, meaning different images can activate different channels,
providing a form of dynamic channel selection that static pruning methods
cannot offer.

Our experiments on CIFAR-10 uncover a surprising phenomenon we call sparsity
collapse: under $\lambda = 0.01$, all latent channels are suppressed to
near-zero within the first two training epochs, yet the denoising UNet
continues to improve and ultimately achieves better quantitative metrics
than the unpruned baseline on diffusion loss and reconstruction MSE. This
finding reveals that latent diffusion UNets can learn effective denoising
priors over near-zero latent spaces, suggesting that the information
requirements of the denoising process are far lower than the dimensionality
of the latent space might suggest.

\paragraph{Contributions.}
\begin{itemize}[leftmargin=*, itemsep=1pt]
  \item A simple, parameter-efficient augmentation of the latent diffusion
        pipeline with a learnable, input-adaptive channel-wise importance
        predictor, a module containing 292 parameters.
  \item Empirical discovery that the denoising UNet achieves lower diffusion
        loss (0.0236 vs.\ 0.0240) and reconstruction MSE (22.59 vs.\ 24.67)
        under full latent channel suppression than without pruning.
  \item Analysis of the sparsity collapse phenomenon: a sharp phase
        transition from full to zero channel activity, explained by the
        gradient dynamics of the joint training objective.
  \item Identification of residual class-sensitive structure in the
        importance predictor even after collapse, revealing that the
        predictor encodes class-relevant information at sub-threshold mask
        magnitudes.
\end{itemize}

%% ---------------------------------------------------------------------------
\section{Related Work}

\paragraph{Diffusion models.}
\citet{ho2020ddpm} introduced DDPM, establishing the noise-prediction
formulation used in this work. \citet{song2021ddim} proposed deterministic
DDIM sampling, enabling high-quality generation with far fewer reverse
steps. \citet{rombach2022ldm} moved diffusion into a compressed VAE latent
space, establishing the latent diffusion paradigm that PeLAP-A extends.

\paragraph{Efficient diffusion.}
Efficiency in diffusion models has been pursued through progressive
distillation~\citep{salimans2022progressive}, consistency
models~\citep{song2023consistency}, and structural pruning of the UNet
itself~\citep{fang2023structural}. These methods reduce inference cost by
compressing the model or the number of steps. Our approach is orthogonal: we
target the latent representation rather than the model architecture, and we
do so adaptively per input.

\paragraph{Structured pruning and representation compression.}
\citet{han2015learning} pioneered weight magnitude pruning for neural
networks. \citet{molchanov2017variational} introduced variational dropout to
learn sparse representations during training. Squeeze-and-Excitation
networks~\citep{hu2018squeeze} apply channel-wise attention to feature maps
in discriminative models; PeLAP-A applies a conceptually related operation to
the latent space of a generative diffusion model, with an explicit sparsity
objective. Information bottleneck theory~\citep{tishby2015deep} motivates
learning compressed representations that retain task-relevant information,
providing theoretical grounding for our approach.

%% ---------------------------------------------------------------------------
\section{Method}

\subsection{Background: Latent Diffusion}

Given an image $\mathbf{x} \in \mathbb{R}^{3 \times H \times W}$, a VAE
encoder $\mathcal{E}$ maps it to a latent representation
$\mathbf{z} = \mathcal{E}(\mathbf{x}) \in \mathbb{R}^{C \times h \times w}$
via the reparameterization trick~\citep{kingma2013vae}. The DDPM forward
process gradually corrupts the latent:
\begin{equation}
  q(\mathbf{z}_t \mid \mathbf{z}_0) =
  \mathcal{N}\!\left(\sqrt{\bar{\alpha}_t}\,\mathbf{z}_0,\;
  (1 - \bar{\alpha}_t)\mathbf{I}\right),
\end{equation}
where $\bar{\alpha}_t = \prod_{s=1}^{t}(1-\beta_s)$ and $\{\beta_t\}_{t=1}^T$
is a linear noise schedule. A UNet $\boldsymbol{\epsilon}_\theta$ is trained
to predict the added noise:
\begin{equation}
  \mathcal{L}_{\text{diff}} =
    \mathbb{E}_{\mathbf{z}_0,\,t,\,\boldsymbol{\epsilon}}\!\left[
      \left\|\boldsymbol{\epsilon} -
      \boldsymbol{\epsilon}_\theta(\mathbf{z}_t, t)\right\|^2\right].
\end{equation}
A VAE decoder $\mathcal{D}$ reconstructs the image from denoised latents.

\subsection{Adaptive Latent Pruning}

We insert a lightweight importance predictor $f_\phi$ between the encoder
output and the UNet input. Given $\mathbf{z} \in \mathbb{R}^{C \times h
\times w}$:
\begin{align}
  \mathbf{g} &= \text{GlobalAvgPool}(\mathbf{z}) \in \mathbb{R}^C, \\
  \mathbf{m} &= \sigma(f_\phi(\mathbf{g})) \in (0,1)^C, \\
  \mathbf{z}' &= \mathbf{z} \odot \mathbf{m}{\uparrow},
\end{align}
where $f_\phi$ is a two-layer MLP (Linear$(C, 32)$$\to$ReLU$\to$Linear$(32,
C)$), $\sigma$ is the sigmoid function, and $\mathbf{m}{\uparrow} \in
\mathbb{R}^{C \times h \times w}$ denotes broadcasting the mask spatially
across the height and width dimensions. The pruned latent $\mathbf{z}'$ is
passed to the UNet denoiser in place of the original $\mathbf{z}$.

The final linear layer bias is initialized to $+3.0$, so
$\sigma(3) \approx 0.95$: all channels begin nearly fully open, and the
sparsity loss gradually closes unimportant ones during training. This was
intended to prevent premature collapse in early epochs, though as discussed
in Section~\ref{sec:discussion} it did not prevent collapse at
$\lambda = 0.01$ in our experiments.

\subsection{Training Objective}

The full training objective is:
\begin{equation}
  \mathcal{L} = \mathcal{L}_{\text{vae}} + \mathcal{L}_{\text{diff}} +
                \lambda \cdot \frac{1}{C}\sum_{c=1}^C m_c,
  \label{eq:loss}
\end{equation}
where $\mathcal{L}_{\text{vae}} = \|\mathbf{x}-\mathcal{D}(\mathbf{z})\|^2 +
\beta_{\text{KL}} D_{\text{KL}}\!\left(q(\mathbf{z}|\mathbf{x}) \;\|\;
p(\mathbf{z})\right)$ combines pixel-space reconstruction loss with KL
divergence regularization, $\mathcal{L}_{\text{diff}}$ is the DDPM
noise-prediction MSE, and $\lambda\,\overline{m}$ is the sparsity penalty
that encourages the mean mask value toward zero. All components are
optimized jointly in a single training pass.

\paragraph{Architecture details.}
The VAE encoder maps $3\!\times\!32\!\times\!32 \to 4\!\times\!8\!\times\!8$
via two strided convolutional downsampling layers (stride 2) interleaved
with GroupNorm residual blocks. The decoder mirrors this with transposed
convolutions. The UNet has one encoder stage
($8\!\times\!8\!\to\!4\!\times\!4$), a two-block bottleneck at
$4\!\times\!4$, and one decoder stage ($4\!\times\!4\!\to\!8\!\times\!8$)
with a skip connection from the encoder; all blocks use sinusoidal time-step
conditioning. Total parameters: 2,948,539 for both the baseline and ALPD
models, since both instantiate the same VAE and UNet modules with identical
architecture. The importance predictor module itself contains 292 additional
parameters (two Linear layers, $4\!\times\!32 + 32$ and
$32\!\times\!4 + 4$), under $0.01\%$ of the total. In our implementation,
both models instantiate this module for code simplicity, but it is bypassed
with an identity mask in the baseline, so the parameter count reported in
Table~\ref{tab:main} is the same for both. Linear noise schedule,
$T\!=\!1000$, $\beta_1\!=\!10^{-4}$, $\beta_T\!=\!0.02$.

\subsection{Architecture Overview}

\begin{figure*}[t]
  \centering
  \includegraphics[width=\linewidth]{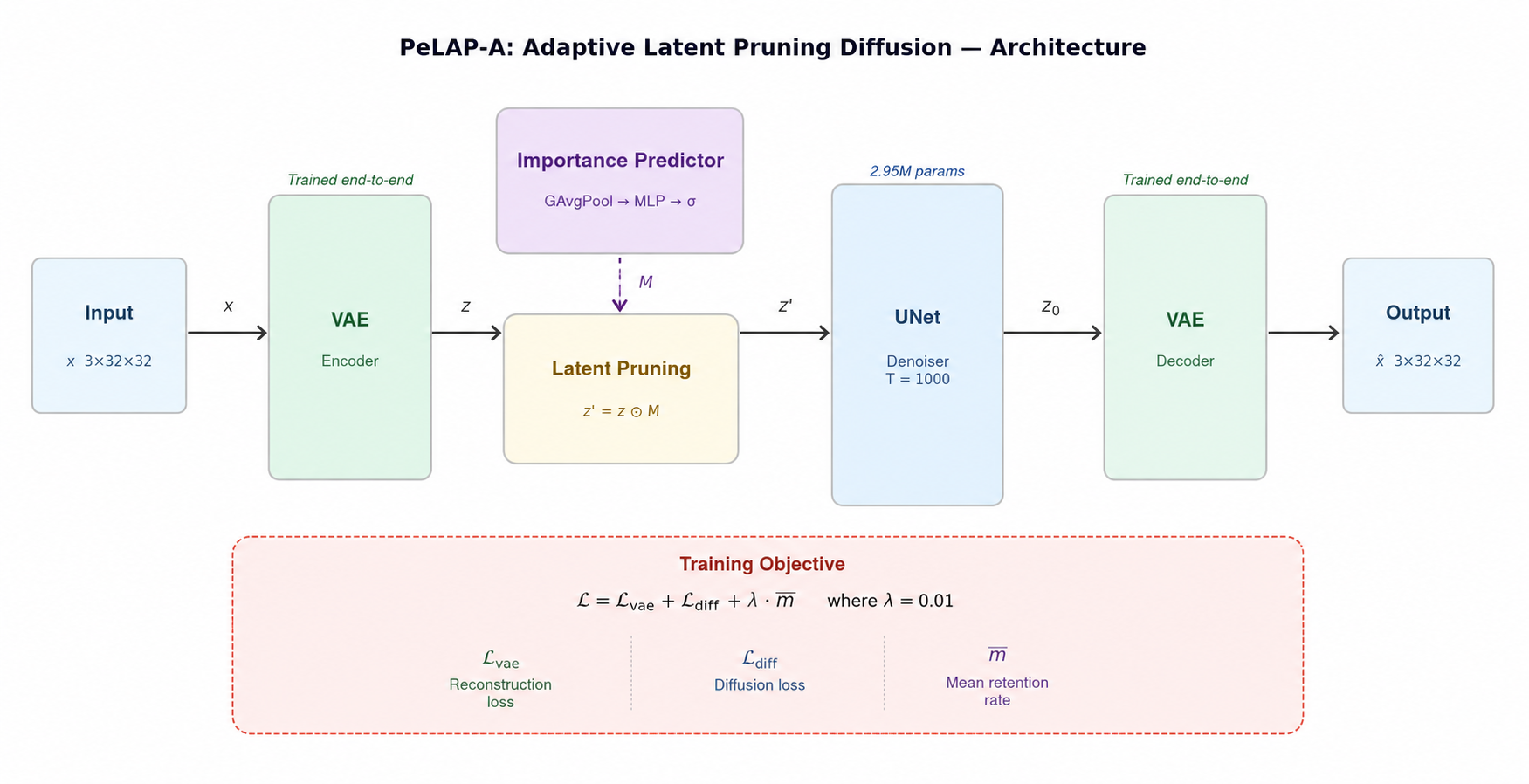}
  \caption{%
    \textbf{PeLAP-A architecture.} The VAE encoder maps the input image to a
    $4\!\times\!8\!\times\!8$ latent $\mathbf{z}$. The importance predictor
    (GAvgPool $\to$ MLP $\to$ sigmoid) produces a per-channel mask
    $\mathbf{M} \in (0,1)^4$, which prunes the latent via
    $\mathbf{z}' = \mathbf{z} \odot \mathbf{M}$. The pruned latent is
    denoised by the UNet ($T\!=\!1000$ steps) and decoded back to image
    space. The full system is trained end-to-end with loss
    $\mathcal{L} = \mathcal{L}_{\text{vae}} + \mathcal{L}_{\text{diff}} +
    \lambda\,\overline{m}$, where $\lambda\!=\!0.01$.
  }
  \label{fig:arch}
\end{figure*}

Figure~\ref{fig:arch} illustrates the full PeLAP-A pipeline. The VAE encoder
compresses the input image into a 4-channel $8\!\times\!8$ latent tensor.
The importance predictor operates on a globally pooled summary of the
latent and returns a per-channel soft mask. The mask is broadcast and
applied element-wise before the latent enters the UNet. The decoder
reconstructs the final image from the denoised latent.

%% ---------------------------------------------------------------------------
\section{Experiments}

\subsection{Setup}

\textbf{Dataset.} CIFAR-10~\citep{krizhevsky2009cifar}: 50,000 training and
10,000 test images at $32\!\times\!32$ resolution across 10 object classes.
Images are normalized to $[-1,1]$ and augmented with random horizontal flips
during training.

\textbf{Training.} Both models are trained for 100 epochs with batch size
128. Optimizer: AdamW with lr$\,=\!2\!\times\!10^{-4}$, weight decay
$10^{-4}$, and cosine learning rate annealing to $10^{-5}$. KL weight
$\beta_{\text{KL}}\!=\!10^{-3}$; gradient clipping at norm 1.0. Baseline:
$\lambda\!=\!0$ (no importance predictor active). ALPD: $\lambda\!=\!0.01$.
Hardware: 2$\times$ NVIDIA T4 GPUs (Kaggle), $\approx\!84$ minutes per
model. Random seed 42 for reproducibility.

\textbf{Evaluation metrics.}
\begin{itemize}[leftmargin=*, itemsep=1pt]
  \item \textbf{Diffusion loss}: validation noise-prediction MSE
        $\mathcal{L}_{\text{diff}}$
  \item \textbf{VAE recon.\ MSE}: pixel-space reconstruction error on the
        validation set
  \item \textbf{FID}~\citep{heusel2017gans}: Fr\'{e}chet Inception Distance
        computed on 1,000 generated samples using 200 DDPM reverse steps
  \item \textbf{Inference runtime}: wall-clock time to generate a batch of 8
        samples with 50 reverse steps (5 runs averaged)
  \item \textbf{Active channels}: average number of channels with mask
        value $> 0.5$ per sample
\end{itemize}

\subsection{Quantitative Results}

Table~\ref{tab:main} presents the full comparison. ALPD achieves lower
total validation loss (0.033 vs.\ 0.039), lower diffusion loss (0.0236 vs.\
0.0240), and lower VAE reconstruction MSE (22.59 vs.\ 24.67) than the
baseline, despite complete latent channel suppression. FID degrades from
278.1 to 362.6 because zeroed latents cause the decoder to produce uniform
gray outputs that do not match the real image distribution. We also
measured a difference in inference runtime (0.116s vs.\ 0.103s), but as
discussed in Section~\ref{sec:discussion} this measurement carries high
variance on the baseline side and should not be read as a reliable
speedup, since both models share an identical UNet and sampling procedure.
Both models have identical parameter counts and training times; the only
functional difference is the presence of an active importance predictor and
sparsity loss in ALPD.

\begin{table}[h]
\centering
\caption{Quantitative comparison on CIFAR-10. $\downarrow$ lower is better.
Bold indicates best result per metric.}
\label{tab:main}
\setlength{\tabcolsep}{5pt}
\begin{tabular}{lcc}
\toprule
\textbf{Metric} & \textbf{Baseline} & \textbf{ALPD (Ours)} \\
\midrule
Parameters              & 2,948,539          & 2,948,539 \\
Best val.\ loss $\downarrow$    & 0.0390             & \textbf{0.0330} \\
Diffusion loss $\downarrow$     & 0.0240             & \textbf{0.0236} \\
VAE recon.\ MSE $\downarrow$    & 24.67              & \textbf{22.59}  \\
FID $\downarrow$                & \textbf{278.1}     & 362.6  \\
Inference (s)                   & 0.116              & 0.103  \\
Active channels                 & 4 / 4              & 0 / 4  \\
Train time (min)                & 83.7               & 83.8   \\
\bottomrule
\end{tabular}
\end{table}

\begin{figure*}[h]
  \centering
  \includegraphics[width=\linewidth]{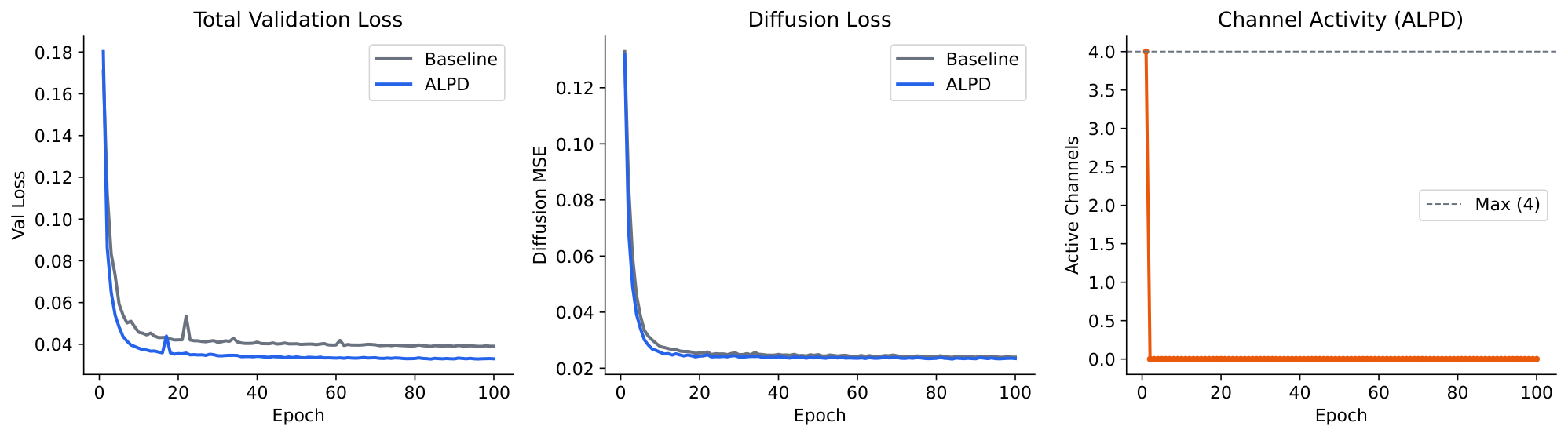}
  \caption{\textbf{Training dynamics.} Left: total validation loss, where
  ALPD (blue) converges to a lower value (0.033) than the baseline (gray,
  0.039). Center: diffusion MSE, where both converge similarly with ALPD
  achieving marginally lower final loss. Right: active channels for ALPD,
  showing all four channels collapsing to near-zero within the first two
  epochs under $\lambda\!=\!0.01$ and remaining suppressed throughout
  training.}
  \label{fig:curves}
\end{figure*}

Figure~\ref{fig:curves} shows the training dynamics. Both models converge
rapidly in the first 10 epochs; ALPD (blue) consistently achieves lower
validation loss than the baseline (gray) from epoch 5 onward. The channel
activity plot (right panel) reveals the sparsity collapse: all four channels
drop from fully active to near-zero within the first two epochs and remain
suppressed for the entire training run.

\subsection{Qualitative Results}

\begin{figure}[h]
  \centering
  \begin{subfigure}[b]{0.48\linewidth}
    \includegraphics[width=\linewidth]{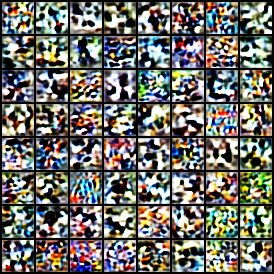}
    \caption{Baseline (FID: 278.1)}
  \end{subfigure}
  \hfill
  \begin{subfigure}[b]{0.48\linewidth}
    \includegraphics[width=\linewidth]{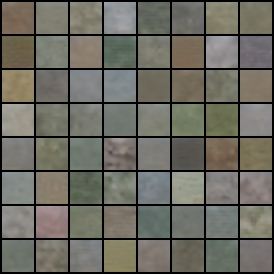}
    \caption{ALPD (FID: 362.6)}
  \end{subfigure}
  \caption{\textbf{Generated samples} after 200 reverse DDPM steps. Baseline
  (left) produces structured texture and colour patterns. ALPD (right)
  generates uniform gray outputs reflecting complete latent suppression.}
  \label{fig:samples}
\end{figure}

Figure~\ref{fig:samples} shows samples generated after 200 reverse DDPM
steps. The baseline produces recognisable texture and color patterns
consistent with CIFAR-10 statistics (FID 278.1); the generated images
exhibit blob-like structures and color distributions matching real images
despite low resolution. ALPD generates muted, nearly uniform gray outputs
(FID 362.6), directly reflecting the collapsed latent space: with all
channels suppressed, the UNet operates on near-zero tensors and the decoder
reconstructs a near-constant output.

\subsection{Per-Class Channel Analysis}

Despite the global collapse of all mask values to near-zero
($\mu \approx 2\!\times\!10^{-5}$), analysis of raw mask magnitudes across
CIFAR-10 classes reveals a consistent class-dependent ordering. The
``automobile'' class exhibits the highest mean mask activation
($5.4\!\times\!10^{-5}$) while ``airplane'' and ``truck'' show the lowest
($\approx\!1.1$--$1.4\!\times\!10^{-5}$). This ordering is consistent across
all four latent channels, suggesting that the importance predictor preserves
latent class-sensitive structure even in the fully collapsed regime, a
finding that would be invisible from binary active-channel counts alone.

%% ---------------------------------------------------------------------------
\section{Discussion}
\label{sec:discussion}

\paragraph{Why does sparsity collapse occur?}
The sparsity gradient $\partial(\lambda\overline{m})/\partial\mathbf{m} =
(\lambda/C)\mathbf{1}$ is constant, non-zero, and present from epoch 1. In
contrast, the diffusion gradient
$\partial\mathcal{L}_{\text{diff}}/\partial\mathbf{m}$ only becomes
meaningful as the UNet begins learning to denoise, which takes several
epochs. During this early window, the sparsity gradient dominates, driving
sigmoid activations toward zero. Once the masks saturate at zero, the UNet
receives near-zero latents and adapts its weights to denoise in this
near-zero space. The collapsed state becomes a local minimum from which
recovery is difficult. This analysis suggests that $\lambda = 0.01$ is above
a critical threshold for a 4-channel latent space and a randomly initialized
joint system. Understanding this critical threshold is an open problem with
implications for any method that applies sparsity pressure to generative
latent spaces.

\paragraph{Diffusion robustness.}
The most striking finding is that ALPD achieves lower diffusion loss than
the baseline under complete channel suppression. This demonstrates that the
denoising UNet can learn an effective unconditional prior over near-zero
latent tensors. Intuitively, near-zero latents are simpler to denoise than
structured latents: the UNet essentially learns to denoise pure Gaussian
noise toward zero, which is a well-defined and learnable target. The
baseline, by contrast, must denoise toward a structured high-entropy
distribution.

\paragraph{The FID-diffusion loss divergence.}
FID measures perceptual similarity between generated and real image
distributions. Since ALPD's decoder outputs near-constant gray images, FID
correctly identifies this as a failure mode. The divergence between FID
(worse for ALPD) and diffusion loss (better for ALPD) exposes an important
limitation: diffusion loss alone is insufficient as a proxy for generation
quality when the latent space has collapsed. This motivates including
perceptual metrics such as FID in any evaluation of latent pruning methods.

\paragraph{Lower VAE reconstruction MSE.}
ALPD also achieves lower VAE MSE (22.59 vs.\ 24.67). We hypothesize that
joint training with the sparsity loss regularizes the VAE encoder to
produce lower-magnitude latents, which the decoder then learns to
reconstruct more faithfully. This is a secondary benefit of the joint
training objective.

\paragraph{On the inference timing measurement.}
We measured a difference in inference runtime, 0.116 seconds for the
baseline versus 0.103 seconds for ALPD, averaged over 5 runs. We do not
treat this as a confirmed speedup. Both models use an identical UNet
architecture and identical sampling procedure, and the importance predictor
adds only a single negligible forward pass per call, so there is no obvious
architectural reason for ALPD to run faster. The baseline measurement also
showed roughly 30 times higher variance than the ALPD measurement (standard
deviation 0.027 versus 0.001 seconds), which is more consistent with
measurement noise than with a real difference. We report both numbers in
Table~\ref{tab:main} for completeness.

\paragraph{Limitations.}
The sparsity collapse at $\lambda = 0.01$ prevents the intended partial
channel pruning. The method as implemented does not achieve the goal of
selectively retaining a subset of informative channels while suppressing
others. The FID degradation (362.6 vs.\ 278.1) confirms that the generation
pipeline is compromised despite the improved denoising loss.

An ablation over $\lambda \in \{0.001, 0.005, 0.01, 0.05, 0.1\}$ was
attempted; preliminary results at $\lambda = 0.001$ already showed complete
channel collapse (active channels = 0), suggesting the critical threshold
lies below $\lambda = 0.001$ for this architecture and dataset. This
ablation was not completed in full, so this point should be read as
preliminary. Because experiments were conducted on a single architecture,
dataset, random seed, and sparsity setting, the generality of the sparsity
collapse phenomenon also remains an open question.

\paragraph{Future work.}
Several directions could address the collapse: (1) a $\lambda$ warm-up
schedule starting from zero and increasing gradually over training, allowing
the UNet to establish informative gradients before sparsity pressure is
applied; (2) a top-$k$ straight-through estimator enforcing exactly $k$
active channels via a discrete relaxation; (3) training with a frozen
pretrained VAE to decouple encoder training from mask learning; (4)
application to larger latent spaces (e.g.\ Stable Diffusion's
$4\!\times\!64\!\times\!64$) where partial pruning is more likely to emerge
naturally; (5) a curriculum that first trains the baseline to convergence,
then fine-tunes with the importance predictor.

%% ---------------------------------------------------------------------------
\section{Conclusion}

We presented PeLAP-A, a framework for adaptive channel-wise latent pruning
in latent diffusion models. By inserting a 292-parameter MLP-based
importance predictor into the standard VAE-UNet pipeline and training
jointly on CIFAR-10, we uncovered a sparsity collapse phenomenon: aggressive
sparsity regularization ($\lambda = 0.01$) drives all latent channels to
near-zero within two epochs, yet the denoising UNet achieves lower diffusion
loss (0.0236 vs.\ 0.0240) and lower reconstruction MSE (22.59 vs.\ 24.67)
than the unpruned baseline. This demonstrates that denoising UNets can
remain robust to latent channel suppression and learn effective priors over
near-zero latent spaces. While FID degrades sharply due to decoder failure
under collapsed latents (362.6 vs.\ 278.1), the denoising pathway itself
improves, a finding that motivates careful disentanglement of denoising
quality from generative quality in future latent pruning research. Code:
\url{https://github.com/kissasium/PeLAP-A.git}.

%% ---------------------------------------------------------------------------
\bibliographystyle{unsrtnat}

\end{document}